%% file: rank_distribution.tex
\documentclass[]{interact}

\usepackage{epstopdf}
\usepackage[caption=false]{subfig}

\usepackage[natbibapa,nodoi]{apacite} 
\theoremstyle{plain}

\theoremstyle{definition}

\theoremstyle{remark}

\usepackage{xcolor}
\usepackage{etoolbox} 
\usepackage{graphicx}
\usepackage{hyperref}
\usepackage{pdflscape} 

\newcommand{\loglikelihood}{{\cal L}}

\newtoggle{anonymous}
\togglefalse{anonymous}

\begin{document}

   \title{The exponential distribution of the order of demonstrative, numeral, adjective and noun.} 


\iftoggle{anonymous}{}
{ 
  \author{
  \name{Ramon Ferrer-i-Cancho\thanks{CONTACT Ramon Ferrer-i-Cancho. Email: rferrericancho@cs.upc.edu}}
  \affil{Quantitative, Mathematical and Computational Linguistics Research Group, Departament de Ci\`encies de la Computaci\'o, Universitat Polit\`ecnica de Catalunya (UPC), Barcelona, Catalonia (Spain). ORCiD: 0000-0002-7820-923X}
  }
}

\maketitle

\begin{abstract}
The frequency of the preferred order for a noun phrase formed by demonstrative, numeral, adjective and noun has received significant attention over the last two decades. We investigate the actual distribution of the 24 possible orders. There is no consensus on whether it is well-fitted by an exponential or a power law distribution. We find that an exponential distribution is a much better model. This finding and other circumstances where an exponential-like distribution is found challenge the view that power-law distributions, e.g., Zipf's law for word frequencies, are inevitable. We also investigate which of two exponential distributions gives a better fit: an exponential model where the 24 orders have non-zero probability (a geometric distribution truncated at rank 24) or an exponential model where the number of orders that can have non-zero probability is variable (a right-truncated geometric distribution). When consistency and generalizability are prioritized, we find higher support for the exponential model where all 24 orders have non-zero probability. 
These findings strongly suggest that there is no hard constraint on word order variation and then unattested orders merely result from undersampling, consistently with Cysouw's view.
\end{abstract}


\section{Introduction}

\label{sec:introduction}


The frequency of the preferred order of a noun phrase formed by demonstrative (D), numeral (N), adjective (A) and noun (n) has received substantial attention over the last 20 years \citep{Cinque2005a, Cysouw2010a, Dryer2018a, Medeiros2018a, Martin2020a}. 
The 24 possible orders are listed in the 1st column of Table \ref{tbl:frequency}.
Researchers have attempted to shed light on the actual variation of frequency among the 24 possible orders with some degree of precision \citep{Cinque2005a,Cysouw2010a,Dryer2018a}. 
A key research question is why not all possible orders are attested. Certain researchers have attributed this to the existence of hard constraints limiting word order variation \citep{Cinque2005a, Cinque2013a,Medeiros2016a, Medeiros2018a}. A hard constraint is a constraint that makes certain word orders impossible. A soft constraint is one that reduces the probability of certain orders, but never to zero probability. In contrast, Cysouw hypothesized that all orders are {\em a priori} possible but some are not attested due to undersampling \citep{Cysouw2010a}.

\begin{figure}[h]
\begin{center}
\includegraphics[width=\textwidth]{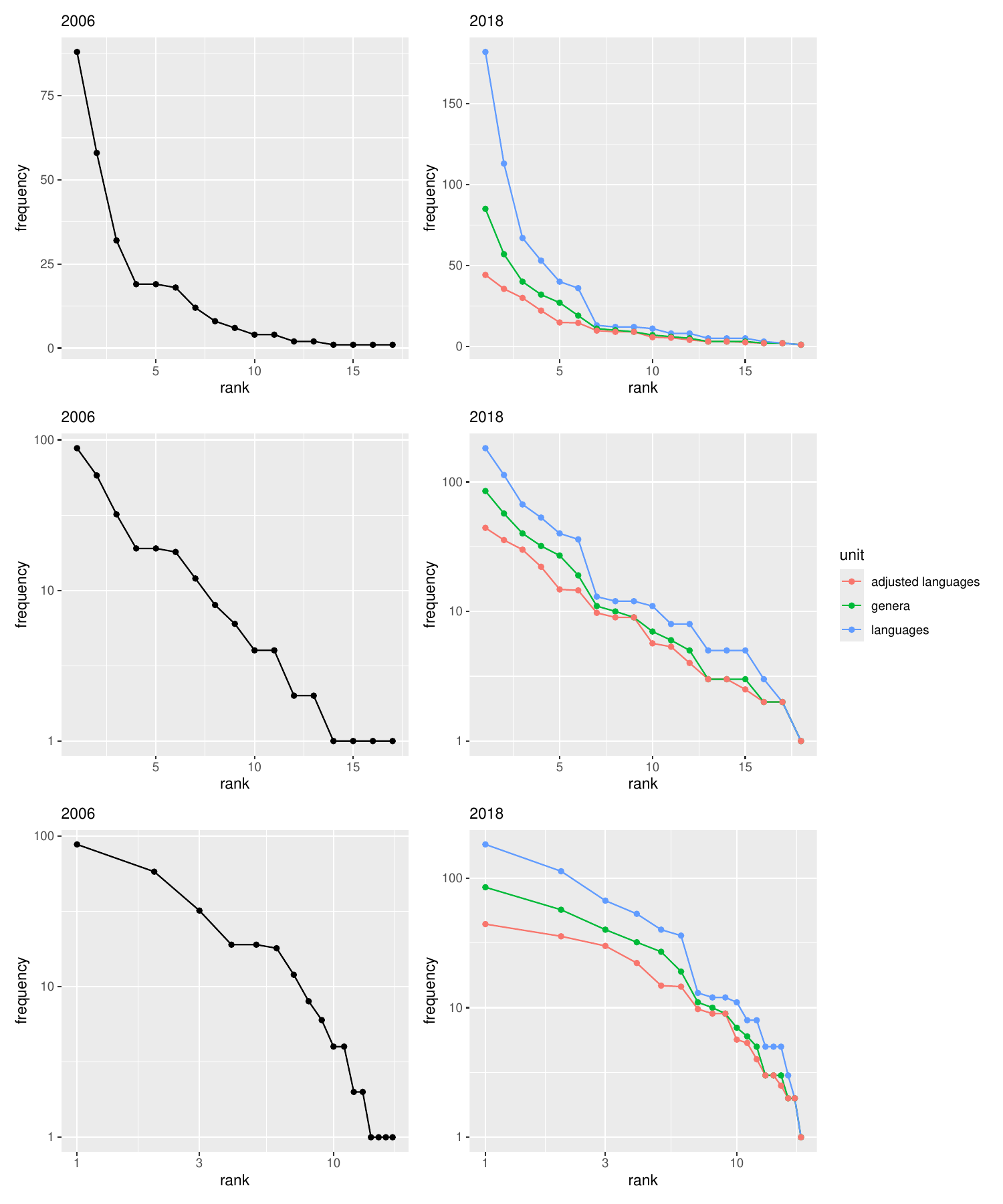}
\end{center}
\caption{\label{fig:rank_distribution}
The frequency of a word order of rank $r$, $f(r)$, in three distinct scales: normal (top), linear-log (middle) and log-log (bottom). Left. Frequency is measured in languages according to \citet{Dryer2006a}.  
Right. Frequency is measured in languages, genera and adjusted number of languages according to \citet{Dryer2018a}.  }
\end{figure}

Concerning the distribution of the rank $r$ of a preferred order (the most frequent order has rank 1, the second most preferred order has rank 2, and so on), shown in Figure \ref{fig:rank_distribution} top,
\citet{Cysouw2010a} proposed an exponential distribution while \citet{Martin2020a} 
assumed a power law distribution. Therefore, another key question is which of the two distributions is more appropriate.
The frequency of rank $r$, i.e. $f(r)$, would follow a power law distribution if $f(r)$ could be approximated by 
\begin{equation}
f(r) = c r^{-\alpha},
\label{eq:power_law_distribution_template}
\end{equation}
where $c$ and $\alpha$ are some positive constants. The parameter $\alpha$ is the so-called exponent of the power law. 
$f(r)$ would follow an exponential distribution if it could be approximated by  
\begin{equation}
f(r) = c e^{-\beta r},
\label{eq:exponential_distribution}
\end{equation}
where $c$ and $\beta$ are some positive constants. \footnote{Notice that $\alpha, \beta \geq 0$ by the definition of $r$.}

In the remainder of the article, we address the two key questions above.
We will show that an exponential distribution is a much better model than a power law distribution and will discuss its implications for the existence of hard constraints in the noun phrase and the meaningfulness of linguistic laws. 
 
\section{Methods}

\subsection{Data}

\label{subsec:data}

\begin{table}[h]
\caption{\label{tbl:frequency}
The frequency of the preferred order of the 24 orders of noun (n), adjective (A), numeral (N) and demonstrative (D) according to \citet{Dryer2018a} and \citet{Dryer2006a}. Orders are sorted according to their frequency in languages.}
\begin{center}
\begin{tabular}{lllll}
\hline
      & \multicolumn{3}{c}{\citet{Dryer2018a}} & \citet{Dryer2006a} \\
\cmidrule(lr){2-4} \cmidrule(lr){5-5}        
order & languages & genera & adjusted languages & frequency \\
\hline
\input tables/frequency_table
\hline
\end{tabular}
\end{center}
\end{table}

Table \ref{tbl:frequency} shows the frequency of the preferred order of demonstrative, numeral, adjective and noun in our dataset. We borrow these frequencies from two datasets: \citet{Dryer2006a} and \citet{Dryer2018a}. 
\citet[Appendix]{Cysouw2010a} displays the frequencies in \citet{Dryer2006a}.
In \citet{Dryer2018a}, the frequency of each preferred order is available in languages, genera and adjusted number of languages. A genus ({\em genera} in plural) is a notion introduced by \citet{Dryer1989a} to refer to a group of closely related languages that is, essentially, an intermediate genetic classification between a language family and a language. Such a classification was applied notably in the World Atlas of Linguistic Structures. \footnote{\url{https://wals.info/languoid/genealogy} 
A genus represents languages that are clearly related, but not so distantly that their relationship is controversial. It typically corresponds to a time depth of about 3,000–4,000 years of divergence, hence it is deeper than subfamilies (like Romance) but shallower than broad families (like Indo-European). For instance,
\begin{itemize}
\item
Indo-European family $\rightarrow$ Romance genus $\rightarrow$ Catalan, French, Italian, ...
\item
Niger-Congo family $\rightarrow$ Bantu genus $\rightarrow$ Swahili, Zulu, Xhosa,...
\end{itemize}
Languages that are genetically or geographically close may not be statistically independent \citep{Winter2021a}. 
}
The adjusted number of languages is an adjustment to the number of languages introduced by \citet{Dryer2018a} to control for geographic proximity or genetic relatedness.  
Frequencies in languages or genera are integer numbers while the frequency in adjusted number of languages is non-integer.

The frequencies above differ from frequencies used in traditional corpus studies in the sense that they are calculated on the preferred orders of languages. Consider the frequency of nAND in languages, that is 182 according to \citet{Dryer2018a} (Table \ref{tbl:frequency}). This frequency is the number of languages that prefer the order nAND (and not any of the other 23 possibilities) for a noun phrase consisting of a demonstrative, a numeral, an adjective and a noun. It is not the frequency of that order in a corpus, although one would expect nAND to be the most frequent in a corpus in each of the 182 languages.
 
For convenience, we define $F_x$ as  
\begin{equation*}
F_x = \sum_{r=1}^{r_{max}}f(r)r^x,
\end{equation*}
where $r_{max}$ is the maximum rank in the sample. Thus, $r_{max}$ is also the number of attested orders, i.e. the number of distinct orders observed in the sample.
The quantity $F_0$ is just the total frequency and is also the sample size. The quantity $F_1$ is the sum of the ranks in our sample (every rank contributes to the sum with as many summands as its frequency). 
The average rank is then
\begin{equation*}
\left< r \right> = F_1/F_0.
\end{equation*}

\begin{table}[h]
\caption{\label{tbl:elementary_summary}
Summary of the elementary statistical properties of each dataset: the total frequency ($F_0$), the average frequency rank ($\left< r \right>$) and the maximum frequency rank in the sample ($r_{max}$). }
\begin{center}
\begin{tabular}{llllllll}
\hline
dataset & unit & $F_0$ & $\left< r \right>$ & $r_{max}$ \\
\hline
\input tables/elementary_summary_table
\hline
\end{tabular}
\end{center}
\end{table}

Table \ref{tbl:elementary_summary} summarizes the elementary statistical properties of the datasets. Within the 2018 dataset, the sample size ($F_0$) reduces while the average rank increases as one moves from languages, to genera and then to adjusted number of languages.

\subsection{The models}

The term power law is used both for continuous and discrete random variables \citep{Debowski2020a,Stumpf2012a,Newman2004a,Conrad2004a,Naranan1998}. 
Similarly, the term exponential distribution is used both for continuous and discrete random variables \citep{Broido2019a,Clauset2009a,Newman2004a,Ferrer2004b}. In these contexts, the term power law is used to refer to a distribution that approximates a power-law function. This is why the term power-law-like distribution is often used. Similarly, the term exponential is used to refer to a distribution that approximates an exponential function. Analogously, one can also use the term exponential-like distribution. 
However, here our random variable, i.e. rank, is discrete. \citet{Martin2020a} stated that the distribution is a power law but did not specify the actual form of the distribution. Our first task is to translate informal terminology into specific discrete distributions \citep{Wimmer1999a,Johnson2005a}.

As there cannot be more than $N = 24$ orders, here we are interested in right-truncated models for $p(r)$, namely models that give $p(r) = 0$ for $r > N$.
Among these models, we are interested in models with early right-truncation, namely models that have a parameter $R \leq N$ such that $p(r) = 0$ when $r < 1$ or $r > R$ and $p(r) > 0$ for $1 \leq r \leq R$.
A power-law-like distribution (Equation \ref{eq:power_law_distribution_template}) can be specified as a zeta distribution, namely \citep[664-665]{Wimmer1999a}
\begin{equation}
p(r) = c r^{-\alpha},
\label{eq:power_law_distribution}
\end{equation}
where the normalization factor is $c = 1/\zeta(\alpha)$. In turn, $\zeta(\alpha)$ is the Riemann zeta function, i.e. 
\begin{equation}
\zeta(\alpha) = \sum_{r=1}^{\infty} r^{-\alpha}.
\end{equation}
A zeta distribution is not an adequate power-law model for our setting because that model predicts $p(r) > 0$ for any finite $r$ such that $r > N$. 
For this reason, we consider instead a right-truncated zeta distribution with two parameters, $\alpha$ and $R$, such that the normalization factor becomes \citep[577-578]{Wimmer1999a}
\begin{equation}
c = \frac{1}{H(\alpha,R)}.
\label{eq:normalization_factor_right_truncated_power_law}
\end{equation}
In turn, $H(R,\alpha)$ is the generalized harmonic number in power $\alpha$ 
i.e. 
\begin{equation*}
H(\alpha,R) = \sum_{r=1}^{R} r^{-\alpha}.
\end{equation*} 

An exponential-like distribution (Equation \ref{eq:exponential_distribution}) can be specified as a geometric distribution, i.e.
\begin{equation}
p(r) = c(1-q)^{r - 1},
\label{eq:geometric_distribution}
\end{equation}
where $c$ is a normalization factor and $q \in (0, 1)$ is a parameter. 
The untruncated geometric distribution is obtained with $c=q$ and then $q$ is the only parameter. An adequate version in our setting is the 2-parameter right-truncated geometric distribution, where (Appendix \ref{app:right_truncated_geometric})
\begin{equation}
c = \frac{q}{1 - (1 - q)^R}.
\label{eq:normalization_factor_right_truncated_geometric}
\end{equation}
Technically, the geometric distribution is the discrete analog of the exponential distribution, which is usually defined on continuous random variables \citep[p. 210]{Johnson2005a} (see Appendix \ref{app:exponential_distribution} for the relationship between Equation \ref{eq:exponential_distribution} and \ref{eq:geometric_distribution}). 

In this article, we use the following ensemble of models (the nickname of the model is followed by its definition):
\begin{itemize}
\item 
{\em Zeta 2.} The 2-parameter right-truncated zeta distribution (Equation \ref{eq:power_law_distribution} with $c$ defined by Equation \ref{eq:normalization_factor_right_truncated_power_law}). The two parameters are $\alpha$ and $R$.
\item
{\em Zeta 1.} The 1-parameter right-truncated zeta distribution that is obtained by setting $R = N$ in the Zeta 2 model. The only parameter is $\alpha$.
\item
{\em Geometric 2.} The 2-parameter right-truncated geometric (Equation \ref{eq:geometric_distribution} with $c$ defined by Equation \ref{eq:normalization_factor_right_truncated_geometric}). The two parameters are $q$ and $R$.
\item
{\em Geometric 1.}  A 1-parameter right-truncated geometric that is obtained by setting $R = N$ in the Geometric 2 model. The only parameter is $q$. 
\end{itemize}
Table \ref{tbl:models} summarizes the mathematical definition of each model and its parameters. Recall that we have excluded from the ensemble popular 1-parameter models such as the (untruncated) geometric model or the zeta distribution because $r$ cannot be larger than $N=24$. 

\begin{table}[h]
\caption{\label{tbl:models} The ensemble of models. For each model, we show its definition, the free parameters and the theoretical constraints on the parameters. } 
\begin{center}
\begin{tabular}{llll}
\hline
model & definition & parameters & constraints \\
\hline
Zeta 1 & \(\displaystyle p(r) = \left\{
                                        \begin{array}{ll} 
                                        \frac{1}{H(N, \alpha)}r^{-\alpha} & \text{if~} r \in [1, N]\\
                                        0 & \text{if~} r \notin [1, N] \\
                                        \end{array}
                                     \right. 
              \) 
                 & $\alpha$ & $0 \leq \alpha$ \\
Zeta 2 & \(\displaystyle p(r) = \left\{
                                        \begin{array}{ll} 
                                        \frac{1}{H(R, \alpha)}r^{-\alpha} & \text{if~} r \in [1, R]\\
                                        0 & \text{if~} r \notin [1, R] \\
                                        \end{array}
                                     \right. 
              \) 
                 & $\alpha$, $R$ & $0 \leq \alpha$, $1 \leq R \leq N$ \\
Geometric 1 & \(\displaystyle p(r) = \left\{
                                        \begin{array}{ll} 
                                        \frac{q}{1 - (1 - q)^N} (1-q)^r & \text{if~} r \in [1, N]\\
                                        0 & \text{if~} r \notin [1, N] \\
                                        \end{array}
                                     \right. 
              \) 
                 & $q$ & $0 < q < 1$ \\
Geometric 2 & \(\displaystyle p(r) = \left\{
                                        \begin{array}{ll} 
                                        \frac{q}{1 - (1 - q)^R} (1-q)^r & \text{if~} r \in [1, R]\\
                                        0 & \text{if~} r \notin [1, R] \\
                                        \end{array}
                                     \right. 
              \) 
                 & $q$, $R$ & $0 < q < 1$, $1 \leq R \leq N$ \\
\hline                 
\end{tabular}
\end{center}
\end{table}

\subsection{Visual diagnostic}

\label{subsec:visual_diagnostic}

Consider a plot with $p(r)$ on the $y$-axis and $r$ on the $x$-axis (Figure \ref{fig:rank_distribution}).
A preliminary conclusion about the functional dependence between $p(r)$ and $r$ can be obtained by taking logarithms on one of the axes or both. 
If $r$ followed a power law (Equation \ref{eq:power_law_distribution}), $\log p(r)$ would be a linear function of $\log r$ since
\begin{equation*}
\log p(r) = -\alpha \log r + \log c.
\end{equation*}
Then the plot in double logarithmic scale should show a straight line with a negative slope $-\alpha$. 
If $r$ followed a geometric distribution (Equation \ref{eq:geometric_distribution}), $\log p(r)$ would be a linear function of $r$ since
\begin{equation*}
\log p(r) = (r - 1) \log (1 - q) + \log c.
\end{equation*}
Then the plot in linear scale for the $x$-axis and logarithmic scale for the $y$ axis should show a straight line with a negative slope $\log (1-q)$. 

\subsection{Model selection}
 
We use information-theoretic model selection to find the best model in the ensemble.  
We use the corrected Akaike Information criterion ($AIC_c$) and the Bayesian Information Criterion ($BIC$), that are defined as \citep{Burnham2002a,Wagenmakers2004a}
\begin{eqnarray}
AIC_c & = & -2\loglikelihood + 2 K \frac{F_0}{F_0 - K - 1} \nonumber \\
BIC   & = & -2\loglikelihood + K \log F_0, \label{eq:BIC}
\end{eqnarray}
where $\loglikelihood$ is the maximum log-likelihood of the parameters of the model, and $K$ is the number of parameters of the model.
The best model is the one that minimizes a criterion.


We define $\Delta_i(x)$ as the difference in a score $x$ between the $i$-th model and the best model. 
Thus, $\Delta_i(BIC)$ is the difference in $BIC$ between the $i$-th model and the best model.

We define $w_i(x)$, the weight of model $i$ according to a score $x$, as
\begin{equation*}
w_i(x) = \frac{\exp\left(-\frac{1}{2}\Delta_i(x)\right)}{\sum_j \exp\left(-\frac{1}{2}\Delta_j(x)\right)}.
\end{equation*}
$w_i(AIC_c)$, the $AIC$ weight, estimates the probability that model $i$ is the true model of the ensemble \citep{Wagenmakers2004a,Burnham2002a}. 
$w_i(BIC)$, the $BIC$ weight, estimates the probability that model $i$ is the quasi-true model in the ensemble \citep[p. 297]{Burnham2002a}.
The evidence of model $i$ over model $j$ with respect to some score $x$ is defined as the ratio $\frac{w_i(x)}{w_j(x)}$.
For $AIC$ and $BIC$, one has \citep{Wagenmakers2004a}
\begin{eqnarray*}
\frac{w_i(AIC)}{w_j(AIC)} & = & \frac{L_i}{L_j}exp(K_j-K_i)\\
\frac{w_i(BIC)}{w_j(BIC)} & = & \frac{L_i}{L_j}n^{\frac{1}{2}(K_j-K_i)},
\end{eqnarray*} 
where $L_i$, $K_i$, are the likelihood and the number of parameters of model $i$. For $AIC_c$, it is easy to see that
\begin{eqnarray*}
\frac{w_i(AIC_c)}{w_j(AIC_c)} & = & \frac{L_i}{L_j} exp\left[F_0 \left(\frac{K_j}{F_0-K_j-1} - \frac{K_i}{F_0-K_i-1} \right) \right].\\
\end{eqnarray*} 

An obvious difference between $AIC_c$ and $BIC$ is that $BIC$ introduces a stronger penalty for lack of parsimony than $AIC_c$ \citep{Wagenmakers2004a}. $BIC$ is more useful in selecting a correct model, which is our primary aim, while $AIC$ is more appropriate in finding the best model for predicting future observations \citep{Chakrabarti2011a}. $AIC$ is not a consistent score in the sense that
as the number of observations $F_0$ tends to infinity, the probability that $AIC$ recovers a true low-dimensional model does not converge to 1 \citep{Wagenmakers2004a}.

We revisit the derivation of $\loglikelihood$ for the Zeta 2 model
\iftoggle{anonymous}{{\bf(author, year)}}
{\citep{Baixeries2012c}}
and extend it to the Geometric 2 model. 
The likelihood of a sample of ranks $\{r_1,...,r_i,...r_{F_0}\}$ can be expressed as
\begin{equation}
L = \prod_{i=1}^{F_0} p(r_i)
\label{eq:likelihood}
\end{equation} 
and then $\loglikelihood = \log L$ can be expressed as 
\begin{equation}
\loglikelihood = \sum_{r=1}^{r_{max}} f(r) \log p(r).
\label{eq:loglikelihood}
\end{equation}
For the Zeta 2 model (Equation \ref{eq:power_law_distribution} with Equation \ref{eq:normalization_factor_right_truncated_power_law}), the log-likelihood is
\begin{equation}
\loglikelihood = - \alpha \sum_{r=1}^{r_{max}}f(r) \log r - F_0 \log H(R, \alpha).
\label{eq:loglikelihood_power_2}
\end{equation}

For the Geometric 2 model, we derive $\loglikelihood$ by plugging Equation \ref{eq:geometric_distribution} (with $c$ defined as in Equation \ref{eq:normalization_factor_right_truncated_geometric}) into the general definition of log-likelihood in Equation \ref{eq:loglikelihood}. After some algebra, one obtains
\begin{equation}
\loglikelihood = F_0 \log \frac{q}{1 - (1 - q)^R} + (F_1 - F_0)\log(1-q).
\label{eq:loglikelihood_geometric_2}
\end{equation}
To obtain the best parameters of a model by maximum likelihood, we proceed as follows. If the model has one parameter, we use one-dimensional optimization. For the Geometric 1 model, the parameter $q$ is optimized in the interval $(0,1)$. For the Zeta 1 model, the parameter $\alpha$ is optimized in the interval $[0,10^6]$. For the models that have two parameters, we note that maximum $\loglikelihood$ requires $R = r_{max}$ (Appendix \ref{app:optimal_truncation}). Therefore, to maximise $\loglikelihood$, we set parameter $R$ to $r_{max}$ and optimize the other parameter following the same procedure as for the 1-parameter models.

\section{Results}

\subsection{Visual diagnostic}

We examine the look of plots of $p(r)$ as a function of $r$ (Figure \ref{fig:rank_distribution}) following the reasoning in Section \ref{subsec:visual_diagnostic}. In normal scale, a decreasing curve is observed (Figure \ref{fig:rank_distribution} top) and it is difficult to determine whether the distribution is power-law-like or exponential-like. 
In logarithmic scale only for $p(r)$, a straight line with negative slope appears, which is compatible with a geometric distribution (Figure \ref{fig:rank_distribution} middle). 
In double logarithmic scale, $p(r)$ curves down, which is incompatible with a power law function (Figure \ref{fig:rank_distribution} bottom). If a distribution curves down in double logarithmic scale, that implies that the probability decay is faster than expected for a power law distribution. To sum up, an exponential distribution is a better candidate than a power law.

\begin{figure}[h]
\begin{center}
\includegraphics[width=\textwidth]{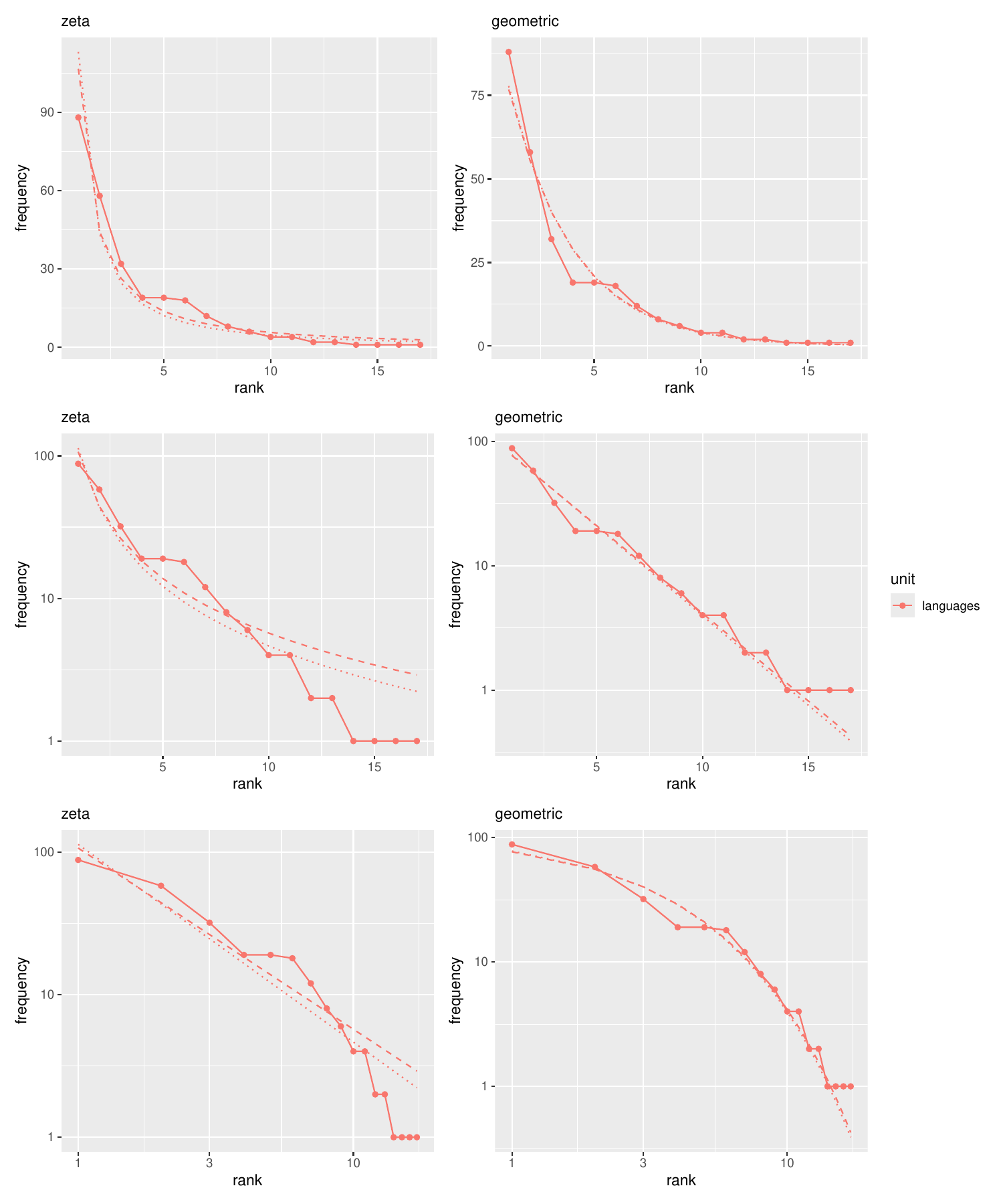}
\end{center}
\caption{\label{fig:rank_distribution_power_geometric_2006}
$f(r)$, the frequency of a word order of rank $r$ in \citet{Dryer2006a}, in three distinct scales: normal (top), linear-log (middle) and log-log (bottom). The solid line is the real curve and the discontinuous lines are the expected value of $f(r)$, that is $\mathbb{E}[f(r)] = F_0 p(r)$, where $p(r)$ is defined by the best fit of a model (Table \ref{tbl:best_parameters_summary}). Left. Real curves versus the best fit of the Zeta 1 model (dotted line) and that of Zeta 2 model (dashed line). Right. Real curves versus the best fit of Geometric 1 (dotted line) and that of Geometric 2 (dashed line). }
\end{figure}

\begin{figure}[h]
\begin{center}
\includegraphics[width=\textwidth]{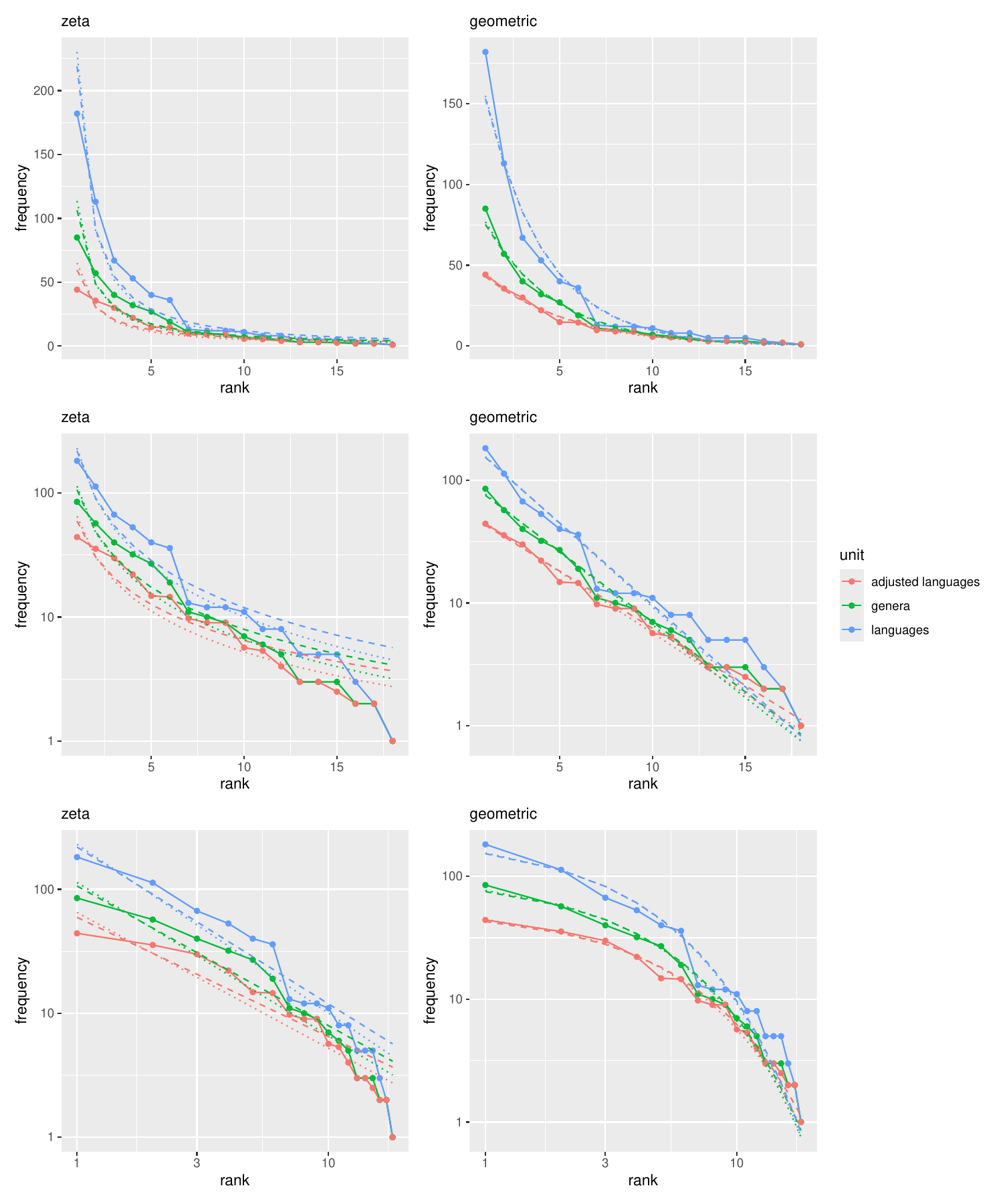}
\end{center}
\caption{\label{fig:rank_distribution_power_geometric_2018}
$f(r)$, the frequency of a word order of rank $r$ in \citet{Dryer2018a}. The format is the same as in Figure \ref{fig:rank_distribution_power_geometric_2006}. }
\end{figure}

\subsection{Model selection}

\label{sec:model_selection_results}

\begin{table}[h]
\caption{\label{tbl:best_parameters_summary}
Summary of the best parameters. For every dataset, frequency unit and model, we show the value of the parameters that maximize 
$\loglikelihood$ ($R$ is the number of non-zero probability ranks, $\alpha$ is the exponent of the power-law models and $q$ is a parameter of the geometric models).}

\begin{center}
\begin{tabular}{llllll}
\hline
dataset & unit & model & $R$ & $\alpha$ & $q$ \\
\hline
\input tables/best_parameters_table 
\hline
\end{tabular}
\end{center}
\end{table}

\begin{landscape} 
\begin{table}[h]
\caption{\label{tbl:model_selection_summary}
Summary of the model selection. For every dataset, frequency unit and model, we show the log-likelihood ($\loglikelihood$), the corrected Akaike Information Criterion ($AIC_c$), the $AIC_c$ difference ($\Delta(AIC_c)$), the $AIC_c$ weight ($w(AIC_c)$), the Bayesian Information Criterion ($BIC$), the BIC difference ($\Delta(BIC)$) and the $BIC$ weight ($w(BIC)$). }
\begin{center}
\begin{tabular}{lllllllllll}
\hline
dataset & unit & model & ${\cal L}$ & $AIC_c$ & $\Delta(AIC_c)$ & $w(AIC_c)$ & $BIC$ & $\Delta(BIC)$ & $w(BIC)$\\
\hline
\input tables/model_selection_table 
\hline
\end{tabular}
\end{center}
\end{table}
\end{landscape}

A stronger conclusion on the best model is obtained by inspecting the $AIC/BIC$ scores (Table \ref{tbl:model_selection_summary}). First, the geometric distribution models give lower $AIC_c$ or $BIC$ than the power law distribution models. The $AIC_c$ weights indicate that the power law model is unlikely to be the true model of the ensemble ($w < 5\cdot 10^{-5}$ for power law models).
Similarly, the $BIC$ weights indicate that the power law model is unlikely to be the quasi-true model of the ensemble ($w < 3\cdot 10^{-5}$ for power law models). Visual inspection confirms it: the theoretical curve of the geometric model is closer to the actual distribution than the theoretical model of the power law (Figures \ref{fig:rank_distribution_power_geometric_2006} and \ref{fig:rank_distribution_power_geometric_2018}). The power-law fails simultaneously by overestimating $p(1)$ and not capturing the faster decay of the actual distribution.
The best values of $q_i$, the value of $q$ of Geometric $i$ model, indicate that $|q_1 - q_2|$ is small but always $q_1 > q_2$ (Table \ref{tbl:best_parameters_summary}). 
In addition, $q_i$ is close to $1/\left< r \right>$, the maximum likelihood estimator of $q$ for an untruncated geometric distribution (Equation \ref{eq:geometric_distribution} with $c = q$).

When comparing Geometric 1 and Geometric 2 with information criteria, the results depend on the score. $AIC_c$ provides more support for Geometric 2 while $BIC$ provides more support for Geometric 1. In particular,
\begin{itemize}
\item
{\em $AIC_c$}. Geometric 2 is better than Geometric 1 in terms of $AIC_c$, except in the 2006 dataset, where the $AIC$ of Geometric 2 is just 0.1 nats above Geometric 1. 
We consider the evidence of Geometric 2 over Geometric 1 by means of $w_2(AIC_c)/w_1(AIC_c)$. In the 2006 dataset, Geometric 2 is about as likely to be the true model as Geometric 1 ($w_2(AIC_c)/w_2(AIC_c) \approx 1$). In the 2018 dataset, Geometric 2 is about 2 times more likely to be the quasi-true model than Geometric 1 when frequency is measured in languages, about 3 times more likely when frequency is measured in genera and about 7 times higher when frequency is measured in adjusted number of languages. 
\item
{\em $BIC$}. Geometric 1 is better than Geometric 2 in terms of $BIC$, except for adjusted number languages in the 2018 dataset, where the $BIC$ of Geometric 1 is just 0.6 nats above Geometric 2. However, the sample size is the smallest for adjusted number of languages (Table \ref{tbl:elementary_summary}) which is a hindrance for model selection with $BIC$ \citep[p. 288]{Burnham2002a}; recall the dependence of BIC on sample size in Equation \ref{eq:BIC}. 
We consider the evidence of Geometric 1 over Geometric 2 by means of $w_1(BIC)/w_2(BIC)$.
The $BIC$ weights indicate that Geometric 1 is about 6 times more likely to be the quasi-true model than Geometric 2 in the 2006 dataset ($w_2(BIC)/w_2(BIC) \approx 6$). In the 2018 dataset, Geometric 1 is about 4 times more likely to be the quasi-true model than Geometric 2 when frequency is measured in languages, about 7/3 times more likely when frequency is measured in genera but about 3/4 ``higher'' when frequency is measured in adjusted number of languages.
\end{itemize}

\section{Discussion}

\subsection{Is the distribution of preferred orders in languages exponential or power-law?}

\subsubsection{The best distribution}
\label{subsec:best_distribution}

We have shown that the geometric distribution, a discrete exponential-like distribution, is a much better model than a power law distribution, both according to visual diagnostic and model selection. Our conclusion is consistent with Cysouw's proposal \citep{Cysouw2010a} and at odds with Martin et al.'s assumption of a power law \citep{Martin2020a}. The expectation of a power law has led to misclassify the rank distribution of vocalizations produced by other species as power laws by means of visual diagnostic \citep{Howes-Jones1988a,Dreher1961a,Janik2006a}. In Figure 12 (p. 24), \citet{Howes-Jones1988a} show the frequency of the calls of the warbling vireo ({\em Vireo vilgus vilgus}) as a function of their rank. 
In figure 2 (p. 1800), \citet{Dreher1961a} shows the frequency of the ``tonemes'' produced primarily by dolphins ({\em Tursiops
truncatus}). In both cases, authors conclude that finding a straight line in linear-log scale agrees with Zipf's law for word frequencies. However, we have shown that a straight line in that scale is indeed indicative of an exponential distribution (Section \ref{subsec:visual_diagnostic}). Thus power laws are less ubiquitous than usually believed.  

\subsubsection{The correct exponential distribution}

We have found that the best model depends on the score. According to $AIC_c$, Geometric 2 has more evidence than Geometric 1 (except for the 2006 dataset). According to $BIC$, Geometric 1 has more evidence (except for adjusted number of languages in the 2018 dataset). However, a stronger conclusion can be reached by looking at the ability of the models to generalize in perspective. 
\citet{Cinque2005a} reported only 14 attested orders. Although he did not report the frequencies of each attested on a sample, we can be totally certain that the best fit of Geometric 2 by maximum likelihood would conclude that only $R = 14$ orders have non-zero probability (justification in Appendix \ref{app:optimal_truncation}), which we know it would be wrong because 17 and 18 orders were found later on \citep{Dryer2006a,Dryer2018a}. We also know that the best fit of Geometric 2 to the 2006 dataset, namely $R = 17$ is misleading because $R=18$ for the 2018 dataset (Table \ref{tbl:best_parameters_summary}). Thus, Geometric 2 fails to generalize twice. 

When integrating likelihood into an information theoretic criterion and considering the 2006 dataset again, $AIC_c$ weights conclude that Geometric 1 is as likely as Geometric 2 (recall $w_2(AIC_c)/w_2(AIC_c) \approx 1$) but $BIC$ weights are able to realize that Geometric 1 is more likely to be the best (recall $w_1(BIC_c)/w_2(BIC) \approx 6$). Crucially, $AIC_c$ does not foresee that the best model in the 2006 dataset, Geometric 2, produces a zero likelihood when applied to the 2018 dataset (notice that the application of the best fit of Geometric 2 for the 2006 dataset to the 2018 dataset yields $p(18) = 0$, which produces $L = 0$ when applied to equation \ref{eq:likelihood}). In contrast, the best model for the 2006 dataset according to BIC yields a non-zero likelihood when applied to the 2018 dataset. In sum, BIC catches early the model that generalizes while $AIC_c$ is unable to realize that one of the models overfits the 2006 dataset. Therefore, $AIC_c$ lacks generalizability.

The failure of $AIC_c$ on the 2006 dataset is not surprising given the theoretical properties of $AIC_c$ versus $BIC$ \citep{Wagenmakers2004a}. $AIC$ is not
{\em consistent} in the sense that, as the number of observations ($F_0$ in our notation) grows very large, the probability that $AIC_c$ recovers a true low-dimensional model does not converge to $1$ \citep[p. 357]{Bozdogan1987a}.
$BIC$ is consistent as the number of observations tends to infinity. In our application, that means that if we supplied to $AIC_c$ a dataset comprising all languages on Earth or even all languages that ever existed in our planet, it would not be warranted that $AIC_c$ would recover the right geometric models while $BIC$ would be much closer to finding the right geometric model.
$BIC$ is more useful in selecting a correct model (in this case, discarding an incorrect model) while the AIC is more appropriate in finding the best model for predicting future observations \citep{Chakrabarti2011a}.
In our application, the main goal is to find the right geometric model, not a geometric model that predicts the actual shape of the distribution when future observations are incorporated.
 

Above, we have provided empirical evidence of the failure of $AIC_c$ to find a correct model in our application. However, we wish to highlight the theoretical power of $BIC$ for our research problem and our datasets.
Once we have discarded the power-law models, the model selection reduces to choosing between two geometric models. $BIC$ assumes equal prior probability on each model \citep{Burnham2002a}. That implies assigning initial equal chance to early truncation (Geometric 2) and to late truncation (Geometric 1).
Indeed, such a contest between two nested models, Geometric 1 and Geometric 2, can be seen as a {\em dimensionality guessing problem} (Geometric 2 adds on dimension with respect to Geometric 1) and $BIC$ is a consistent estimator of $K$, the dimensions of the ``true model'' \citep[p. 284]{Burnham2002a}. $BIC$ implicitly assumes that {\em ``truth is of fairly low dimension (e.g., $K = 1-5$) and that the data-generating (true) model is fixed as sample size increases''} \citep[286]{Burnham2002a}. 
Various sources of evidence suggest that the true model's dimensionality is bounded by a small number in our application.
\citet{Dryer2018a} accurately modelled the frequency rank of the orders by means of 5 binary descriptive principles, suggesting that the true number of dimensions of the distribution is bounded above by $K = 5$. Indeed, just $\lfloor \log_2 N \rfloor + 1 = 5$ binary parameters suffice to sort $N$ orders so as to match the desired frequency rank. Previously, \citet{Cysouw2010a} had proposed exponential models for the frequency of orders with 3, 4, or 6 binary parameters. Therefore, we apply BIC under favourable conditions, where the dimensions of truth are bounded. In our application, $BIC$ always provides more evidence for Geometric 1 except for adjusted number of languages in the 2018 dataset, which coincides with the smallest sample (Table \ref{tbl:elementary_summary}). The consistency property of $BIC$ requires a large enough sample (Equation \ref{eq:BIC} and \citet[p. 288]{Burnham2002a}). 
Therefore, our findings so far and the conditions where we are applying $BIC$ suggest that Geometric 1 is a stronger model for the underlying distribution.

\subsection{Are there hard constraints on word order?}

The quick answer to this question is that there is no statistically robust evidence for a hard constraint limiting word order variation in the noun phrase. A detailed discussion follows.

Once the power law models are discarded, the question of the existence of hard constraints reduces to a dimensionality guessing problem, i.e. which of the two geometric models is the best. 
The evidence for Geometric 1 is a challenge for the existence of hard constraints limiting word order variation in languages that would explain why not all 24 possible orders are attested \citep{Cinque2005a, Cinque2013a,Medeiros2016a, Medeiros2018a}. 
If there were no hard constraints (only soft constraints), the best model should be Geometric 1; if strong constraints existed, Geometric 2 (with $R < 24$) should be the best model.
The strength of Geometric 1 is consistent with Cysouw's hypothesis: all orders are {\em a priori} possible but some are not attested due to undersampling \citep{Cysouw2010a}. The challenge of proponents of a hard constraint is two-fold. First, to make a robust proposal, one that does not fall any time that new orders are attested. For instance, the proposal that the 14 orders attested by \citet{Cinque2005a} result from universal grammar or some universal cognitive mechanism, still stands as soft constraint but not as hard constraint \citep{Cinque2005a, Cinque2013a, Medeiros2016a, Medeiros2018a} because so far, 18 orders have been attested \citep{Dryer2018a}. Model selection by means of $AIC_c$ commits overfitting for believing that the current number of attested orders is the true one. In particular, $AIC_c$ leads to parameters for the best model on the data from \citet{Dryer2006a} that do not generalize to the data from \citet{Dryer2018a}. Second, to deal with parsimony. 
Postulating a hard constraint implies a loss of parsimony but is it rewarding enough in terms of proximity to truth? Information criteria address this question and give a compelling answer: when empirical generalizability and theoretical consistency are prioritized (this is the virtue of $BIC$ with respect to $AIC_c$), the absence of a hard constraint (Geometric 1) becomes more likely than its presence (Geometric 2), as shown in Table \ref{tbl:model_selection_summary}).

It is important to understand that the answer to the question on the existence of hard constraints does not follow from choosing BIC because it penalizes for lack of parsimony more strongly than AIC \citep{Wagenmakers2004a}. If we did so, the conclusion that Geometric 1, and thus the absence of hard constraints, would be trivial. The relevant questions, that summarize the discussions above, are the following: 
\begin{enumerate}
\item
Do we want a score for which there is no empirical evidence of contradiction in large enough samples? 
\item
Do we want a score that is theoretically consistent, namely, that as more languages are sampled, the probability that the score chooses the right low-dimensional model tends to one?  
\item
Do we want to use a score that is more useful in selecting a correct model (over being more appropriate in finding the best model for predicting future observations)?
\end{enumerate}
If the answer to all the questions above is YES then
\begin{itemize}
\item
The score is BIC.
\item
The best model is Geometric 1.
\item
The absence of a hard constraint is more likely than its presence.
\item
The use of score with a stronger penalty for lack of parsimony with respect to $AIC$ is a side-effect, not a prior desideratum. 
\end{itemize}  

\subsection{The meaningfulness of linguistic laws}

The abundance of exponential-like distributions has implications for the debate on the meaningfulness of linguistic laws \citep[Box 2]{Semple2021a}. Since Zipf's foundational research \citep{Zipf1949a}, many researchers have cast doubt on the depth and utility of linguistic laws such as Zipf's rank-frequency law, the power law that characterizes the distribution of word ranks \citep{Zipf1949a,Moreno2016a,Mehri2017a}. The recurrent criticism that linguistic laws in the form of power laws are inevitable \citep{Miller1963,Sole2010a} can be falsified by finding patterns across different species and systems that do not conform to the law that is claimed to be inevitable \citep{Semple2021a}. Various sources of evidence demonstrate that Zipf's rank-frequency law is not inevitable \citep{Li2002a}. Here we review evidence from exponential distributions. First, our finding that an exponential distribution yields a better fit to word order ranks than a power-law. Second, the exponential distribution that is found in the order of SOV structures \citep{Cysouw2010a} as well as in part-of-speech tags \citep{Tuzzi2010a}, colors, kinship terms, verbal alternation classes \citep{Ramscar2019a}. Third, the exponential rank distribution in the species mentioned above (Section \ref{subsec:best_distribution}) as well as in ``key signs'' produced by rhesus monkeys \citep[Figure 3, p. 367]{Schleidt1973a}.
Fourth, the exponential distribution is found in other linguistic variables such as the distance between syntactically related words \citep{Ferrer2004b,Petrini2022c} or the length of vocal sequences in primates \citep{Gustison2016a,Girard-Buttoz2022a}.
Finally, in non-linguistic contexts, the projection distances between cortical areas exhibit an exponential distribution \citep{Ercsey2013a} and a double exponential distribution characterizes the average distance traversed by foraging ants \citep{Campos2016a}, just to name a few. In sum, there is no empirical support for the claim that power laws are inevitable, even in a linguistic context

\section{Conclusion}

We have shown that a geometric distribution gives a better fit than a power law distribution to the distribution of preferred orders in the noun phase. This has implications
for various components of the generative-linguistics program and other branches of theoretical linguistics that share the same abstractions, e.g., binary acceptability, 
\footnote{The point is assuming that there are impossible orders; whether acceptable orders vary in degree of acceptability is irrelevant for this point.}
or the same methods. 
On the one hand, our analysis shows that statistical evidence for a hard constraint on word order variation in the noun phrase is lacking. Not observing the 24 orders is not enough to claim a hard constraint. History has demonstrated that the post-hoc arguments built on the current number of attested orders \citep{Cinque2005a, Cinque2013a, Medeiros2016a, Medeiros2018a} have collapsed as more orders have been attested. On the other hand, the finding of an exponential distribution challenges another component of the generative-linguistics program, i.e. the inevitability of power laws such as Zipf's rank-frequency law, and consequently, the lack of interest in explaining their origin \citep{Miller1963}. Interestingly, linguistic laws are absent from the agenda in the continuum between generative linguistics and its opponents. When claiming that linguistic universals are a myth, opponents ignore their existence or neglect them \citep{Evans2009a}. An intriguing question is whether they do it for the same reason as generativists and formal linguists from other traditions. While researchers engage in complex debates about the potential advantage of models with a huge number of parameters \citep{Futrell2025a} as if reality were intrinsically of ultra-high dimension, the examination of elementary exponential distributions with a few parameters reveals that a hard constraint on word order lacks statistical support, as we have seen here, while it sheds light on the structure of short-term memory and the dynamics of incremental sentence processing \citep{Petrini2022c}. Reality may be simple, but researchers may fail to see it. 

\section*{APPENDIX}

\appendix

\section{Exponential versus geometric distribution}

\label{app:exponential_distribution}

The primary goal of this appendix is not to point out that the geometric distribution is the discrete analog of the exponential distribution for the continuous case, which is well-known in the community of mathematics and statistics \citep[p. 2010]{Johnson2005a}. The actual goal is to instruct readers from other backgrounds on the fact a geometric distribution for some random variable $x$ can be expressed as an exponential function in a literal sense, that is involving an expression of the form 
\begin{equation}
p(x) = ... e^{f(x)},
\label{eq:exponential}
\end{equation}
where $f(x)$ is some function of $x$.
One of the targets of this appendix is the use of the term exponential distribution or the assumption of an expression of the form of equation \ref{eq:exponential} both for continuous and discrete variables \citep{Clauset2009a,Newman2004a,Ferrer2004b} neglecting that, in the discrete case, that has a precise equivalent that is the geometric distribution.

The customary definition of an exponential distribution for a discrete random variable, e.g., frequency rank (Equation \ref{eq:exponential_distribution}) and the definition of a geometric distribution (Equation \ref{eq:geometric_distribution}) are indeed equivalent. To see it, notice that 
\begin{eqnarray*}
p(r) & = & c (1-q)^{r - 1} \\
     & = & c e^{(r - 1)\log(1-q) } \\
     & = & c' e^{-\beta r}, 
\end{eqnarray*}
where $c' = c/(1 - q)$ and $\beta = -\log(1-q)$.

\section{The right-truncated geometric distribution}

\label{app:right_truncated_geometric}

Our 2-parameter right-truncated geometric distribution is defined on $r = 1,2,...,R$.
\citet{Wimmer1999a} present a similar but not equivalent 2-parameter right-truncated geometric distribution that is defined on $r = 0,1,2,...,R$.
Our 2-parameter right-truncated geometric distribution is obtained when the normalization factor in Equation \ref{eq:geometric_distribution} is  
$c=1/S(1, R)$, where 
\begin{equation*}
S(1, R) = \sum_{r=1}^{R}(1 - q)^{r - 1}.
\end{equation*}

A compact expression for $S(1, R)$ is easy to obtain. By the self-similarity property of the geometric series,
\begin{equation*}
(1-q)S(1, R) = S(1, R) - 1 + (1-q)^{R}.
\end{equation*}
After some simple algebraic manipulations, one obtains 
\begin{equation*}
S(1, R) = \frac{1 - (1 - q)^{R}}{q}
\end{equation*}
and then 
\begin{equation*}
c = \frac{q}{1 - (1 - q)^{R}}.
\end{equation*}
It is easy to check that 
\begin{equation*}
\lim_{R \rightarrow \infty} c = q,
\end{equation*}
which is the normalization factor of the untruncated geometric distribution. 

\section{The value of $R$ that maximizes log-likelihood}

\label{app:optimal_truncation}

We aim to show that maximum $\loglikelihood$ requires $R = r_{max}$ for the Zeta 2 model and the Geometric 2 model.
First, we show that maximum $\loglikelihood$ requires $R \geq r_{max}$. As these models are such that $p(r) = 0$ for $r > R$, setting $R < r_{max}$ implies that the likelihood of the model is $L=0$ (recall Eq. \ref{eq:likelihood}) and the log-likelihood $\loglikelihood$ goes to $-\infty$.   
Second, we show that the log-likelihood functions of these models are monotonically decreasing functions of $R$ when the other parameter ($\alpha$ or $q$) is constant and thus maximum $\loglikelihood$ requires $R = r_{max}$. Let us assume that $\alpha$ and $q$ are constant. For Zeta 2, we revisit previous arguments 
\iftoggle{anonymous}{{\bf(author, year).}}
{\citep{Baixeries2012c}. }

The only summand in Eq. \ref{eq:loglikelihood_power_2} that depends on $R$ is $-F_0 \log H(R, \alpha)$. 
The recursive definition  
\begin{equation*}
H(R, \alpha) = \left\{ 
               \begin{array}{ll}
               1 & \text{if~} R = 1 \\
               H(R - 1, \alpha) + R^{-\alpha} & \text{if~} R > 1
               \end{array}
               \right.
\end{equation*}
clearly shows that $H(R, \alpha)$ is a monotonically increasing function of $R$ (when $\alpha$ is fixed)
and that $-F_0 \log H(R, \alpha) <0$ since $F_0>0$ and $H(R, \alpha) \geq 1$. Therefore, $\loglikelihood$ is a monotonically decreasing function of $R$ (for constant $\alpha$).
For Geometric 2, we recall the assumption $q \in (0,1)$ and note that the 1st derivative of $\loglikelihood$ (Eq. \ref{eq:loglikelihood_geometric_2}) with respect to $R$ is
\begin{eqnarray*}
\frac{\partial \loglikelihood}{\partial R} & = & -F_0 \frac{\partial \log (1 - (1 - q)^R)}{\partial R} \\
                                           & = & F_0 \frac{(1-q)^R \log(1-q)}{1 - (1-q)^R}.
\end{eqnarray*}
It is easy to see that $\partial \loglikelihood / \partial R <0$ because all terms in the expression are strictly positive except $\log(1-q) < 0$.



\iftoggle{anonymous}{}
{ 
\section*{Acknowledgements}

We are grateful to two anonymous reviewers for their very useful feedback. We are also grateful to D. Dediu for very valuable comments and to G. Ramchand 
for making us aware of Medeiro's research. 
This research is supported by the grant PID2024-155946NB-I00 funded by Ministerio de Ciencia, Innovación y Universidades (MICIU), Agencia Estatal de Investigación (AEI/10.13039/501100011033) and the European Social Fund Plus (ESF+).
This research is also supported by a recognition 2021SGR-Cat (01266 LQMC) from AGAUR (Generalitat de Catalunya).
}

\bibliographystyle{apacite}

\bibliography{../../../../Dropbox/biblio/rferrericancho,../../../../Dropbox/biblio/complex,../../../../Dropbox/biblio/ling,../../../../Dropbox/biblio/cl,../../../../Dropbox/biblio/cs,../../../../Dropbox/biblio/maths}

\end{document}

%% file: tables/frequency_table.tex
nAND & 182 & 85 & 44.17 & 88\\
DNAn & 113 & 57 & 35.56 & 58\\
NnAD & 67 & 27 & 14.54 & 32\\
DnAN & 53 & 40 & 29.95 & 19\\
DNnA & 40 & 32 & 22.12 & 18\\
nADN & 36 & 19 & 14.8 & 19\\
nDAN & 13 & 11 & 9 & 8\\
DnNA & 12 & 10 & 9.75 & 1\\
DAnN & 12 & 7 & 5.34 & 12\\
nNAD & 11 & 9 & 9 & 6\\
nDNA & 8 & 6 & 5.67 & 4\\
NAnD & 8 & 5 & 4 & 2\\
AnND & 5 & 3 & 3 & 1\\
NnDA & 5 & 3 & 3 & 4\\
AnDN & 5 & 3 & 2.5 & 2\\
DANn & 3 & 2 & 2 & 0\\
NDAn & 2 & 2 & 2 & 0\\
nNDA & 1 & 1 & 1 & 1\\
NADn & 0 & 0 & 0 & 0\\
NDnA & 0 & 0 & 0 & 1\\
ADnN & 0 & 0 & 0 & 0\\
ADNn & 0 & 0 & 0 & 0\\
ANDn & 0 & 0 & 0 & 0\\
ANnD & 0 & 0 & 0 & 0\\

%% file: tables/elementary_summary_table.tex
2006 & languages & 276 & 3.5 & 17 \\
2018 & languages & 576 & 3.7 & 18 \\
 & genera & 322 & 4.2 & 18 \\
 & adjusted languages & 217.4 & 4.8 & 18 \\

%% file: tables/best_parameters_table.tex
2006 & languages & Geometric 1 & & & 0.282 \\
 &  & Geometric 2 & 17 & & 0.277 \\
 & & Zeta 1 & & 1.386 \\
 & & Zeta 2 & 17 & 1.271 \\
2018 & languages & Geometric 1 & & & 0.269 \\
 &  & Geometric 2 & 18 & & 0.265 \\
 & & Zeta 1 & & 1.361 \\
 & & Zeta 2 & 18 & 1.264 \\
 & genera & Geometric 1 & & & 0.238 \\
 &  & Geometric 2 & 18 & & 0.231 \\
 & & Zeta 1 & & 1.237 \\
 & & Zeta 2 & 18 & 1.126 \\
 & adjusted & Geometric 1 & & & 0.204 \\
 & languages & Geometric 2 & 18 & & 0.193 \\
 & & Zeta 1 & & 1.095 \\
 & & Zeta 2 & 18 & 0.963 \\

%% file: tables/model_selection_table.tex
2006 & languages & Geometric 1 & -581.5 & 1165 & 0 & 0.518 & 1168.6 & 0 & 0.866 \\
 &  & Geometric 2 & -580.6 & 1165.2 & 0.1 & 0.482 & 1172.4 & 3.7 & 0.134 \\
 & & Zeta 1 & -603.3 & 1208.7 & 43.7 & $1.7\cdot 10^{-10}$ & 1212.3 & 43.7 & $2.9\cdot 10^{-10}$ \\
 & & Zeta 2 & -589.8 & 1183.6 & 18.6 & $4.8\cdot 10^{-5}$ & 1190.8 & 22.2 & $1.3\cdot 10^{-5}$ \\
2018 & languages & Geometric 1 & -1244.8 & 2491.6 & 1.7 & 0.304 & 2496 & 0 & 0.793 \\
 &  & Geometric 2 & -1243 & 2490 & 0 & 0.696 & 2498.7 & 2.7 & 0.207 \\
 & & Zeta 1 & -1278.7 & 2559.4 & 69.4 & $5.9\cdot 10^{-16}$ & 2563.7 & 67.7 & $1.5\cdot 10^{-15}$ \\
 & & Zeta 2 & -1254.5 & 2513.1 & 23.1 & $6.7\cdot 10^{-6}$ & 2521.8 & 25.8 & $2\cdot 10^{-6}$ \\
 & genera & Geometric 1 & -738.5 & 1478.9 & 2.1 & 0.255 & 1482.7 & 0 & 0.691 \\
 &  & Geometric 2 & -736.4 & 1476.8 & 0 & 0.745 & 1484.3 & 1.6 & 0.309 \\
 & & Zeta 1 & -766.3 & 1534.5 & 57.8 & $2.1\cdot 10^{-13}$ & 1538.3 & 55.6 & $5.8\cdot 10^{-13}$ \\
 & & Zeta 2 & -748.6 & 1501.3 & 24.6 & $3.5\cdot 10^{-6}$ & 1508.8 & 26.2 & $1.4\cdot 10^{-6}$ \\
 & adjusted & Geometric 1 & -532.8 & 1067.7 & 3.9 & 0.123 & 1071 & 0.6 & 0.427 \\
 & languages & Geometric 2 & -529.8 & 1063.7 & 0 & 0.877 & 1070.4 & 0 & 0.573 \\
 & & Zeta 1 & -555.6 & 1113.3 & 49.5 & $1.5\cdot 10^{-11}$ & 1116.6 & 46.2 & $5.4\cdot 10^{-11}$ \\
 & & Zeta 2 & -539.7 & 1083.5 & 19.8 & $4.4\cdot 10^{-5}$ & 1090.2 & 19.8 & $2.9\cdot 10^{-5}$ \\